\documentclass[letterpaper, 10 pt, conference]{ieeeconf}  

\IEEEoverridecommandlockouts                              

\usepackage{graphicx}
\usepackage{amsmath} 
\usepackage{subcaption}
\usepackage{amssymb}
\usepackage[linesnumbered,ruled,vlined]{algorithm2e}
\usepackage{bm} 
\usepackage{cite}
\usepackage{hyperref}
\usepackage{booktabs} 
\usepackage{tabularx} 
\usepackage{layouts}
\usepackage{jabbrv}
\usepackage{multirow}

\newcommand{\taskname}[1]{{\small \it #1}}

\title{\LARGE \bf
 LLM-Guided Task- and Affordance-Level Exploration\\in Reinforcement Learning
}

\author{
    Jelle Luijkx\textsuperscript{1*}, Runyu Ma\textsuperscript{1*}, Zlatan Ajanovi{\'c}\textsuperscript{2}, and Jens Kober\textsuperscript{1}
    \thanks{
    \textsuperscript{*} Equal Contribution.
    \textsuperscript{1} Cognitive Robotics, Delft University of Technology, The Netherlands.
    \textsuperscript{2} RWTH Aachen University, Germany.
    }%
}

\begin{document}

\bstctlcite{BSTcontrol}

\maketitle
\thispagestyle{empty}
\pagestyle{empty}

\begin{abstract}

Reinforcement learning (RL) is a promising approach for robotic manipulation, but it can suffer from low sample efficiency and requires extensive exploration of large state-action spaces.
Recent methods leverage the commonsense knowledge and reasoning abilities of large language models (LLMs) to guide exploration toward more meaningful states.
However, LLMs can produce plans that are semantically plausible yet physically infeasible, yielding unreliable behavior.
We introduce LLM-TALE, a framework that uses LLMs' planning to directly steer RL exploration.
LLM-TALE integrates planning at both the task level and the affordance level, improving learning efficiency by directing agents toward semantically meaningful actions.
Unlike prior approaches that assume optimal LLM-generated plans or rewards, LLM-TALE corrects suboptimality online and explores multimodal affordance-level plans without human supervision.
We evaluate LLM-TALE on pick-and-place tasks in standard RL benchmarks, observing improvements in both sample efficiency and success rates over strong baselines.
Real-robot experiments indicate promising zero-shot sim-to-real transfer.
Code and supplementary material are available at \url{https://llm-tale.github.io}.

\end{abstract}

\section{Introduction}
Reinforcement learning (RL)~\cite{sutton2018reinforcement} offers a powerful framework for learning decision-making and control policies in robotics~\cite{kober2013reinforcement} through interaction with the environment.
However, practical deployment is hindered by low sample efficiency.
Training stable manipulation policies requires exhaustive exploration of vast state--action spaces and adequate reward feedback.
This challenge is especially problematic when policies are randomly initialized in long-horizon tasks with sparse rewards.
To overcome this problem, intrinsic motivation methods have been introduced to encourage visiting novel states via intrinsic rewards~\cite{stadie2015incentivizing,bellemare2016unifying,achiam2017surprise,ostrovski2017count,pathak2017curiosity}, but the resulting novelty signals can misalign with task-relevant behavior.
A well-known failure mode for some intrinsic-motivation methods is the ``noisy-TV'' thought experiment: an agent can be drawn to a TV showing white noise, since each frame is novel and effectively unpredictable.

Another line of work improves efficiency by incorporating demonstrations that inject human knowledge into off-policy RL~\cite{nair2018overcoming,nair2020awac,ball2023efficient,vecerik2017leveraging,hu2023imitation, bhaskar2024planrl}, but this relies on costly human demonstrations.

Foundation models such as Llama~3~\cite{huang2024good} and GPT-4~\cite{openai2023gpt4}, trained on vast datasets, offer a scalable alternative.
These models act as approximate knowledge sources~\cite{kambhampati2024llms}, leveraging human-like reasoning to aid robotic manipulation.
Recent work shows that large language models (LLMs) and vision--language models (VLMs) can interpret environmental context and perform task-level reasoning, translating open-language commands into sequences of executable skills~\cite{huang2022language,brohan2023can,zeng2022socratic,huang2023voxposer,liang2023code}.
However, their understanding of the physical world is incomplete and can yield erroneous guidance.
Despite these errors, their approximation of human knowledge can enhance RL training by guiding it at higher levels of abstraction (e.g., task or affordance levels).
Using LLMs to generate dense reward functions~\cite{du2023guiding,kwon2023reward,xie2023text2reward,ma2023eureka} aligns robot behavior with human language, mitigating challenges posed by sparse rewards.
Yet such human-like rewards can induce undesirable behavior, e.g., remaining in high-reward regions without accomplishing the task.
Beyond rewards, recent work~\cite{ma2024explorllm,chen2024rlingua} uses LLMs to generate actions directly, guiding robots toward semantically meaningful regions and shifting the data distribution in off-policy replay buffers.
These exploration-driven actions may be suboptimal initially but improve during policy learning.
Nonetheless, the effectiveness of these methods depends heavily on the quality of LLM-generated actions and, at present, is largely limited to off-policy RL.

\begin{figure*}[ht]
    \centering
    \includegraphics[width=\textwidth]{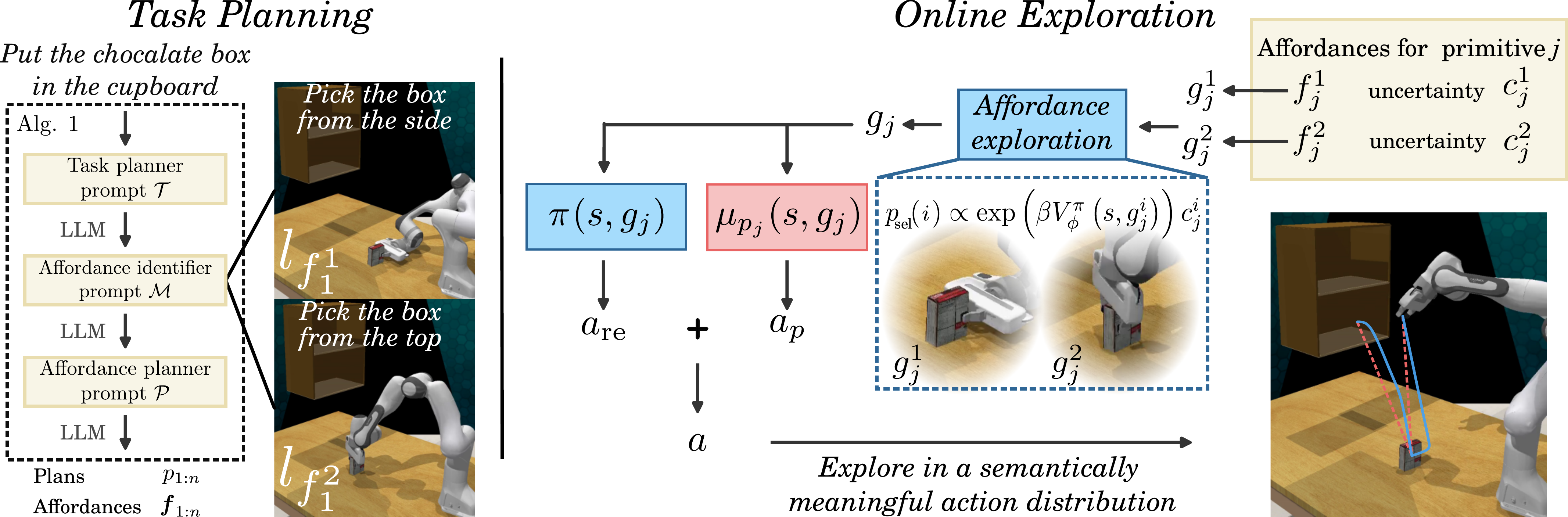}
    \caption{
    LLM-guided Task- and Affordance-Level Exploration (LLM-TALE) uses LLMs to generate task- and affordance-level plans to explore semantically meaningful regions.
    It explores multimodal affordances using goal-conditioned value functions.
    }
    \label{fig:overview}
\end{figure*}

In this paper, we introduce \textit{LLM-guided Task- and Affordance-Level Exploration} (LLM-TALE), a method that steers exploration toward semantically meaningful regions of the state space by leveraging LLM guidance that is more reliable at higher levels of abstraction (i.e., task and affordance).
Because affordance-level actions are often multimodal and some modes are semantically valid yet physically infeasible, our method uses the critic’s value estimates (e.g., Q-values) to explore more promising modes and to avoid fruitless exploration of infeasible actions.

This work introduces LLM-TALE and makes the following contributions:
\begin{enumerate}
\item We propose a hierarchical, LLM-driven planning scheme that generates task-level plans and affordance-level action candidates.
\item We present a goal-conditioned residual RL framework in which goals are derived from LLM-generated affordances, and exploration is guided by intrinsic rewards defined relative to these goals.
\item We introduce critic- and uncertainty-guided affordance-level exploration over LLM-generated proposals, enabling a trade-off between exploration and exploitation across affordance modalities.
\end{enumerate}
These components bias exploration toward semantically meaningful regions of the action space.
Our experiments show high sample efficiency in sparse-reward robotic manipulation for both on- and off-policy RL, while real-world evaluations show promising zero-shot sim-to-real transfer.

\section{Related Work}

\subsection{Exploration in Reinforcement Learning}
Classic exploration methods in reinforcement learning provide intrinsic rewards that encourage visiting novel or uncertain states~\cite{stadie2015incentivizing,bellemare2016unifying,achiam2017surprise,ostrovski2017count}.
While effective at avoiding exploitative actions, such signals can accumulate low-value experience in robotic manipulation.
To address sparse rewards, prior works integrate human demonstrations into replay buffers and guide learning with behavior cloning (BC) losses~\cite{vecerik2017leveraging,nair2018overcoming,nair2020awac}.
RLPD~\cite{ball2023efficient} achieves strong data efficiency by using high update-to-data ratios and ensemble critics to learn from offline data without BC.
IBRL~\cite{hu2023imitation} initializes TD3 agents~\cite{fujimoto2018addressing} from demonstration-trained policies.
However, these approaches require high-quality demonstrations.

\subsection{Reinforcement Learning with Foundation Models}
Foundation models enable task-level reasoning for robotics~\cite{brohan2023can,huang2023grounded,lin2023text2motion,liu2023llm+,chen2023autotamp,dalal2024planseqlearn}.
In RL, they have been used to specify dense reward functions aligned with human language~\cite{du2023guiding,kwon2023reward,xie2023text2reward}, though such rewards can lead to unintended behaviors such as reward hacking (e.g., action repetition without task completion).
Eureka~\cite{ma2023eureka} refines reward code via evolutionary optimization in parallel environments, but the evolutionary loop can be computationally costly.
Recent work~\cite{ma2024explorllm,chen2024rlingua} instead uses foundation models to propose actions directly, guiding robots toward semantically meaningful regions and shifting the data distribution in off-policy replay buffers.
However, such approaches remain dependent on high-quality LLM actions for success.

\subsection{Research Gap}
In contrast to existing methods, LLM-TALE learns an RL policy that operates in a residual action space under LLM guidance to focus exploration on semantically meaningful regions.
By doing so, LLM-TALE induces a more informative online data distribution, without the need for high-quality (human) demonstrations.
Moreover, LLM-TALE addresses LLM reasoning errors online through interaction and explores multimodal affordances identified by the task planner.

\section{Preliminaries}
\subsection{Reinforcement Learning}

We model the environment as a Markov decision process (MDP), i.e., $\mathcal{M} \triangleq (S, A, R, P, \rho_0, \gamma)$, where $S$ and $A$ denote the state and action spaces, respectively. 
The reward function $R$ provides reward $r_t = R(s_t, a_t, s_{t+1})$, and the transition function $P$ defines $P(s_{t+1} | s_t, a_t)$. 
Finally, we have the initial state distribution $\rho_0$ and the discount factor $\gamma$.

The goal of RL is to find an optimal policy $\pi^*$ that maximizes expected cumulative discounted return, with $\pi$ specifying action $a_t \sim \pi(\cdot \mid s_t)$ for each state $s_t$:
\begin{equation}
\pi^* = \arg\max_{\pi} \mathbb{E}_{\pi}\Bigg[ \sum_{t=0}^{T} \gamma^t R(s_t, a_t, s_{t+1}) \Bigg].  
\end{equation}

\begin{figure*}[t]
    \centering
    \includegraphics[width=\linewidth]{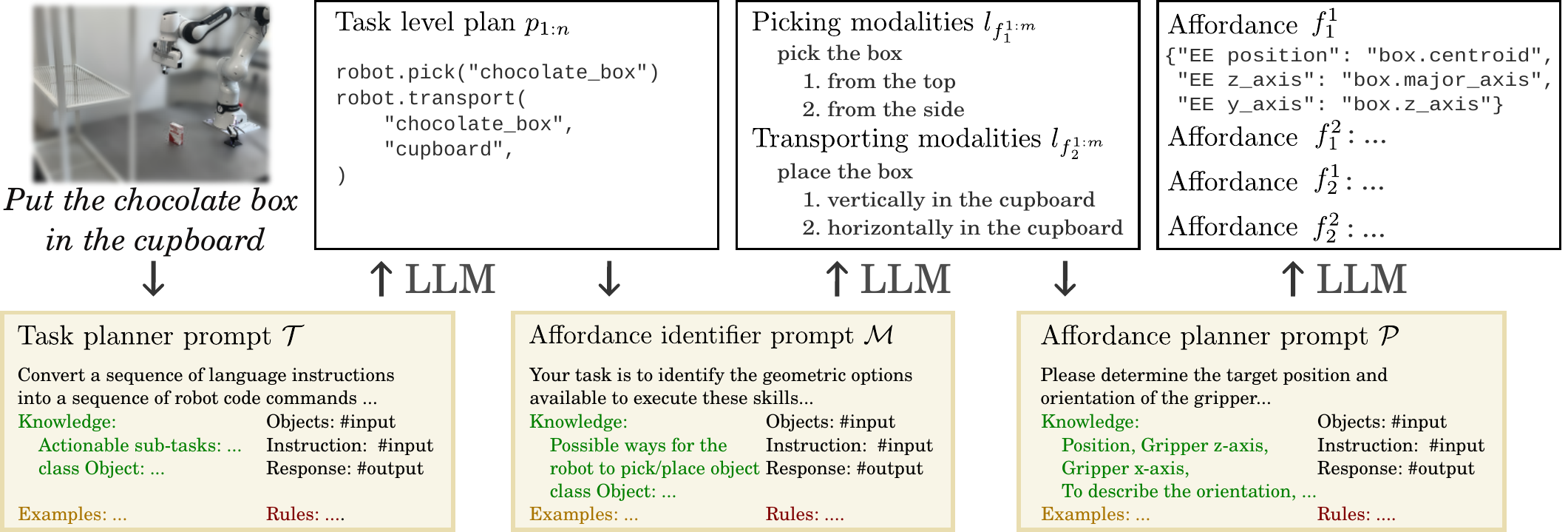}
    \caption{Detailed visualization of the task planner scheme from Alg.~\ref{alg:taskplan}, showing the structure of prompts $\mathcal{T}, \mathcal{M}$ and $\mathcal{P}$.
    }
    \label{fig:taskplanning_detailed}
\end{figure*}

Value functions are ubiquitous in RL algorithms and provide the expected discounted return of a policy given an initial state $V^\pi(s)$ or a state-action pair $Q^\pi(s,a)$.
Because the true value functions are usually unknown, they are often approximated, e.g., with neural networks parameterized by $\phi$. The state value \(V_\phi^{\pi}(s)\) approximates the expected discounted return when following policy \(\pi\) from state \(s\):
\begin{equation}
V^{\pi}(s) = \mathbb{E}_{\pi} \left[ \sum_{t} \gamma^t R(s_t, a_t, s_{t+1}) \Big| \, s_0 = s \right].
\end{equation}
Analogously, $Q^\pi_\phi(s, a)$ estimates the expected discounted return for a state-action pair.
LLM-TALE combines goal-conditioned state or state--action values with an uncertainty metric $c \in (0, 1]$ to trade off exploration and exploitation.

\subsection{Problem Formulation}
We focus on pick-and-place manipulation tasks similar to those described in~\cite{james2020rlbench,mu2021maniskill}.
The environment provides a sparse external reward $R^\mathrm{ex}: S \times A \rightarrow \{0, 1\}$ that indicates task success.
Our goal is to leverage the reasoning ability of LLMs to generate plans that guide exploration toward semantically meaningful regions of the state–action space, while a residual policy compensates for suboptimalities in the generated plans.

\section{Method}

\subsection{Overview of LLM-TALE}
\label{sec:overview}
LLM-TALE consists of a planning phase and a training phase.
Before training, the planning pipeline generates plans at the task and affordance levels.
Task-level planning decomposes a language command into a sequence of \(n\) primitives \(p_{1:n}\) of two types: pick and transport.
Affordance-level plans \(\bm{f}_{1:n}\) describe the identified affordances and their affordance-level plans.
We represent the complete plan as the ordered sequence \(\bm{p} = \big(p_j(\bm{f}_j)\big)_{j=1}^{n}\), where \(p_j \in \{\mathrm{pick}, \mathrm{transport}\}\).
An overview of the planning process is presented in Fig.~\ref{fig:overview}, while detailes are giving in Sec.~\ref{sec:task_planning} and Fig.~\ref{fig:taskplanning_detailed}.

During training, the primitive \(p_j\) translates the affordance-level plan into a goal \(g_j \in SE(3)\) for the end-effector, conditioned on the objects' state \(s^{\text{obj}}\).
These goals are defined relative to an object pose, as in Fig.~\ref{fig:overview}, where the two picking goals correspond to side and top grasps.

A hard-coded PD controller serves as a base policy, \(a_p=\mu_{p_j}(s, g_j)\), and drives the end-effector from the current state toward \(g_j\) along linear trajectories in position and orientation.
The RL agent learns a residual action policy, \(a_{\text{re}}\sim \pi(\cdot | s, g_j)\), as in~\cite{johannink2019residual}.
This residual formulation steers exploration toward semantically meaningful regions online.
The RL agent corrects inaccuracies and suboptimal behavior in LLM controllers, as depicted in Fig.~\ref{fig:overview}.
The executed action is the sum of \(a_p\) and \(a_{\text{re}}\):
\begin{equation}
    a = a_p + a_{\text{re}}.
\end{equation}
To guide exploration toward the goal, we define an intrinsic reward \(r^{\mathrm{in}}\).
This dense reward is primitive-specific and requires no task-specific tuning:
\begin{equation}
    r^{\mathrm{in}} = R^{\mathrm{in}}_j\!\left(s, g_j\right).
\end{equation}
Here, \(R^{\mathrm{in}}_j\) is a dense shaping term computed from pose errors relative to \(g_j\) and the magnitudes of joint velocities.

Our LLM-guided initialization induces a semantically meaningful state distribution.
As a result, the RL agent only refines the policy around the LLM-induced distribution, improving sample efficiency.
The approach is compatible with both on-policy and off-policy algorithms and does not require (human) demonstration data.

\subsection{Task Planning}
\label{sec:task_planning}
This section describes how the LLM converts high-level task instructions into task- and affordance-level plans.
Our pipeline comprises three prompts: a task-level prompt \(\mathcal{T}\), an affordance identifier prompt \(\mathcal{M}\), and an affordance planner prompt \(\mathcal{P}\).
The prompt structure is inspired by \cite{zha2023distilling}, as our work targets similar setups.

As summarized in Alg.~\ref{alg:taskplan} and visualized in Fig.~\ref{fig:taskplanning_detailed}, the task-level prompt \(\mathcal{T}\) translates a high-level task description \(L\) into primitive action descriptions, such as ``\textit{Pick up the cube}'' or ``\textit{Place the box in the cupboard}''.
Querying \(\mathcal{T}\) with \(L\) yields a sequence of Python primitives \(p_{1:n}\) (e.g., \texttt{robot.pick(object)}) for reuse across episodes.

Affordance-level plans are often semantically multimodal, but not every mode is physically feasible.
Consider placing a box in a cupboard (Fig.~\ref{fig:overview}).
The actions of picking and placing the box are ambiguous: it may be grasped from the top or the side and placed horizontally or vertically.
However, placing an open box horizontally may spill its contents, while placing it vertically can make it unreachable for the robot.
This example illustrates that an LLM-generated plan can be semantically valid yet physically infeasible without grounding in the real world.
We use this scenario as a motivating example throughout the paper and refer to it as the \taskname{PutBox} experiment.

Because the LLM lacks the physical understanding required to directly select the best affordance, we introduce an affordance identifier prompt \(\mathcal{M}\) that enumerates semantically distinct affordance plans when language admits multiple modes.
For each primitive \(p_j\), \(\mathcal{M}\) produces language descriptions \(l_{f_j^{1:m}}\) for \(m\) affordance modalities, such as ``\textit{Pick the box from the side}'' or ``\textit{Pick the box from the top}'' (see Alg.~\ref{alg:taskplan}).

The affordance planner prompt \(\mathcal{P}\) then maps each description \(l_{f^i_j}\) to a natural-language affordance \(f_j^i\) that specifies the robot end-effector pose, rather than precise \(SE(3)\) coordinates.
For picking tasks, affordances are specified by three attributes: position, end-effector (EE) z-axis, and EE y-axis (the facing direction of the EE and the direction in which the gripper opens).
This lets the LLM align robot motion with environmental features effectively.
For transport tasks, we use object-centric formulations to accommodate discrepancies between expected and actual picking poses, improving robustness and environmental adaptability.

\begin{algorithm}[t]
\DontPrintSemicolon
\caption{Task Planning}
\label{alg:taskplan}
\SetKwInOut{Prompts}{Prompts}
\KwIn{Task description $L$}
\Prompts{Task-planner prompt $\mathcal{T}$, modality-identifier prompt $\mathcal{M}$, affordance-planning prompt $\mathcal{P}$}
\KwOut{Task-level plans $p_{1:n}$, affordance-level plans $\bm{f}_{1:n}$}

$p_{1:n} \gets \texttt{query\_LLM}\big(\mathcal{T}(L)\big)$\;

\For{$j = 1:n$}{
    $l_{f_j^{1:m}} \gets \texttt{query\_LLM}\big(\mathcal{M}(p_j)\big)$\;
    \For{$i \gets 1$ \KwTo $m$}{
        $f_j^i \gets \texttt{query\_LLM}\big(\mathcal{P}(p_j, l_{f_j^i})\big)$\;
    }
    $\bm{f}_j \gets \{f_j^1,\dots,f_j^m\}$\;
}
\Return $p_{1:n},\;\bm{f}_{1:n}$\;
\end{algorithm}

\begin{algorithm}[t]
\DontPrintSemicolon
\caption{LLM-TALE}
\label{alg:llm-tale}
\SetKwInOut{Parameters}{Parameters}
\KwIn{Task description $L$}
\Parameters{Temperature $\beta$, update rate $\alpha$, minimum uncertainty $c_\mathrm{min}$, number of episodes $K$}
\KwOut{Policy $\pi$}

$p_{1:n},\;\bm{f}_{1:n} \gets \texttt{plan\_task}(L)$ \tcp*{Alg.~\ref{alg:taskplan}}

Initialize residual policy $\pi$, base policies $\mu_{1:n}$, value function $V^\pi_\phi$, uncertainties $\bm{c}_{1:n}$, time step $t \gets 0$\;
\For{$k = 1:K$}{
    $s_t, s^\mathrm{obj}_t \gets \texttt{reset()}$\;
    \ForEach(\tcp*[f]{Primitives}){$p_j \in p_{1:n}$}{
        \ForEach(\tcp*[f]{Affordances}){$f_j^i \in \bm{f}_j$}{
            $g_j^i \gets \texttt{parse\_to\_goal}(s^\mathrm{obj}_t, f_j^i)$\;
            $p_\mathrm{sel}(i) \propto \exp\!\bigl(\beta\, V_\phi^{\pi}(s_t, g_j^i)\bigr)\, c_j^i$\;
        }
        Sample $i \sim p_\mathrm{sel}$ and set $g_j \gets g^{(i)}_j$\;
        $c^{(i)}_j \gets \max\!\bigl((1-\alpha)c^{(i)}_j,\; c_{\min}\bigr)$\;
        \While{$p_j$ not done}{
            $a_t \gets \pi(s_t, g_j) + \mu_j(s_t, g_j)$\;
            $s_{t+1}, s^\mathrm{obj}_{t+1}, r^\mathrm{ex}_t, \mathrm{done} \gets \texttt{step}(a_t)$\;
            $r_t \gets R^\mathrm{in}_j(s_{t+1}, g_j) + r^\mathrm{ex}_t$\;
            Store transition $(s_t, a_t, r_t, s_{t+1}, \mathrm{done})$\;
            Update $\pi$ and $V^\pi_\phi$ with stored transitions\;
            $t \gets t + 1$\;
        }
    }
}
\Return $\pi$
\end{algorithm}

\subsection{Online Exploration}
\label{sec:online_exploration}

\begin{figure}[t]
    \centering
    \includegraphics[width=\linewidth]{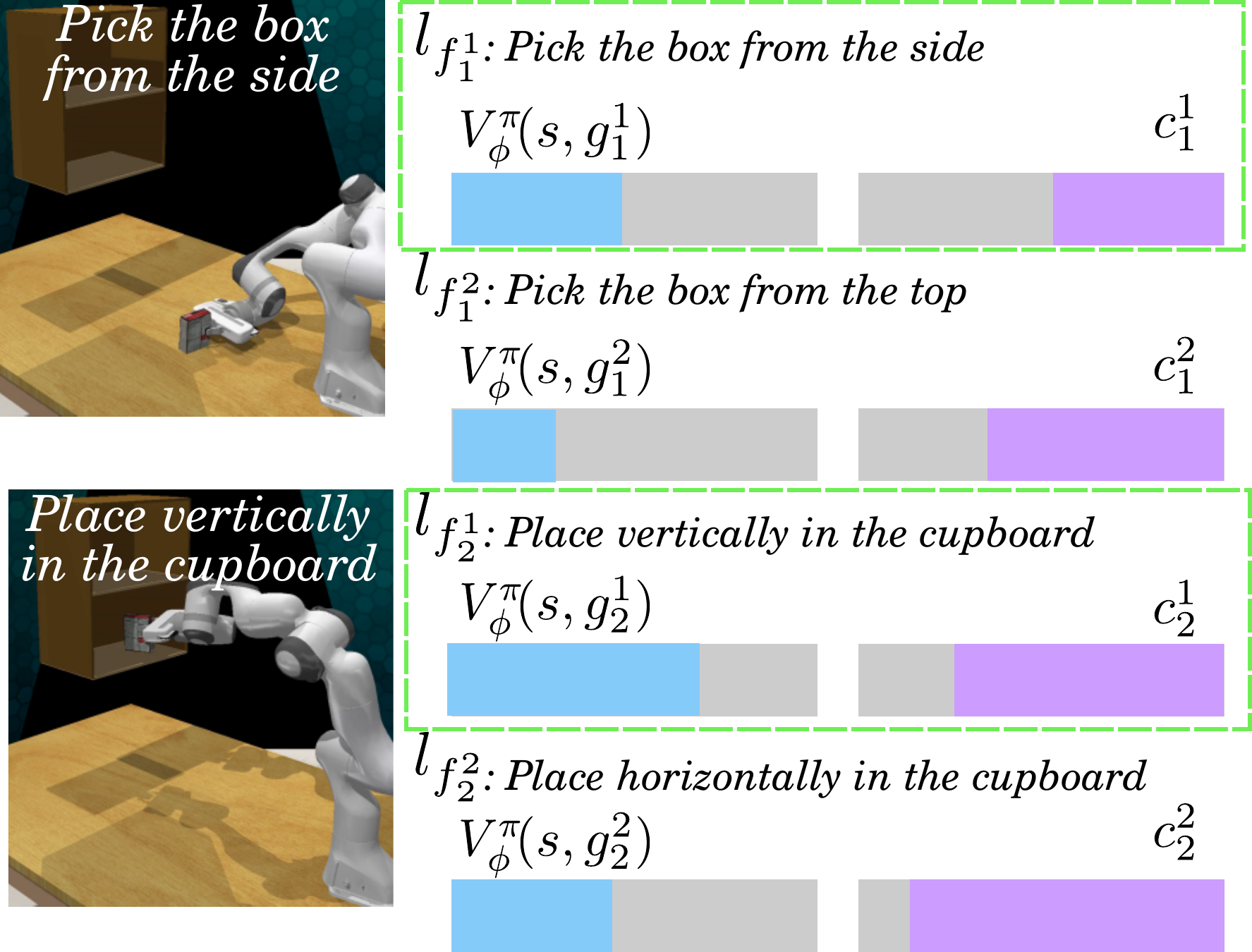}
    \caption{LLM-TALE explores affordance modalities based on value \( V^\pi_\phi(s, g^i_j) \) and uncertainty score \( c^i_j \).}
    \label{fig:multimodal}
\end{figure}

Our method explores affordance multimodality.
During training, affordance plans $f_j^{1:m}$ are parsed into candidate goal poses $g_j^{1:m}$.
For each primitive $p_j$, we score each goal $g_j^i$ with the value function and an uncertainty term, then sample $g_j$ from the resulting distribution.
The value function estimates the expected return from the current state and thus approximates the utility of a goal under the present observation.
For agents with a state-value function, we define the goal selection probabilities as
\begin{equation}
\label{eq:selection}
p_\mathrm{sel}(i) \propto \exp\!\left(\beta\, V_{\phi}^{\pi}\!\left(s, g_j^i\right)\right)\, c_j^i,
\end{equation}
and an analogous expression applies when using a Q-function \(Q_{\phi}(s,a)\).
Here, $\beta>0$ controls the distribution’s sharpness, and $c^{i}_j$ trades off exploration and exploitation.
We then sample $i \sim p_\mathrm{sel}$, set the goal $g_j \gets g^{(i)}_j$, and update the corresponding uncertainty:
\begin{equation}
c^{(i)}_j \gets \max\!\bigl((1-\alpha)c^{(i)}_j,\; c_{\min}\bigr)
\end{equation}
where $\alpha \in (0,1)$ and $c_{\min}$ is a lower bound on uncertainty.
This multimodal affordance exploration strategy is visualized in Fig.~\ref{fig:multimodal}.
The complete procedure of LLM-TALE is described in Alg.~\ref{alg:llm-tale}.

\begin{figure*}
    \centering
    \includegraphics[width=\linewidth]{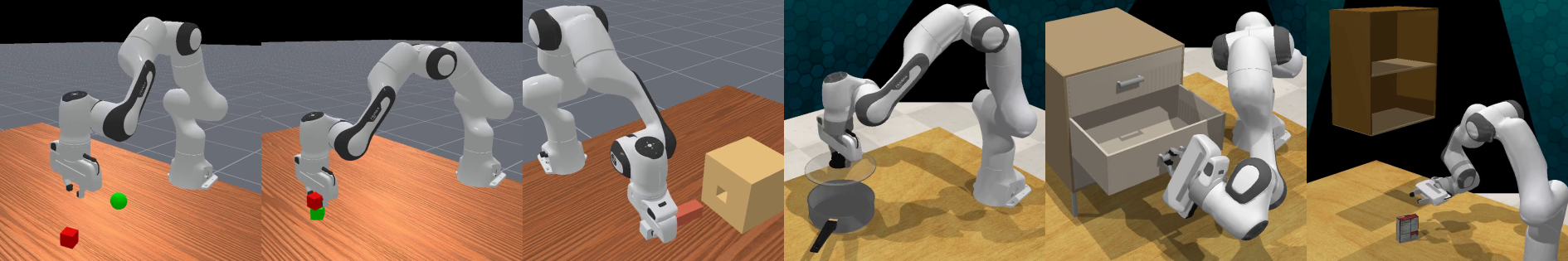}
    \caption{Simulation tasks (left to right): \taskname{PickCube}, \taskname{StackCube}, \taskname{PegInsert}, \taskname{TakeLid}, \taskname{OpenDrawer}, and \taskname{PutBox}.}
    \label{fig:tasks}
\end{figure*}

\begin{figure*}
    \centering
    \includegraphics[width=\linewidth]{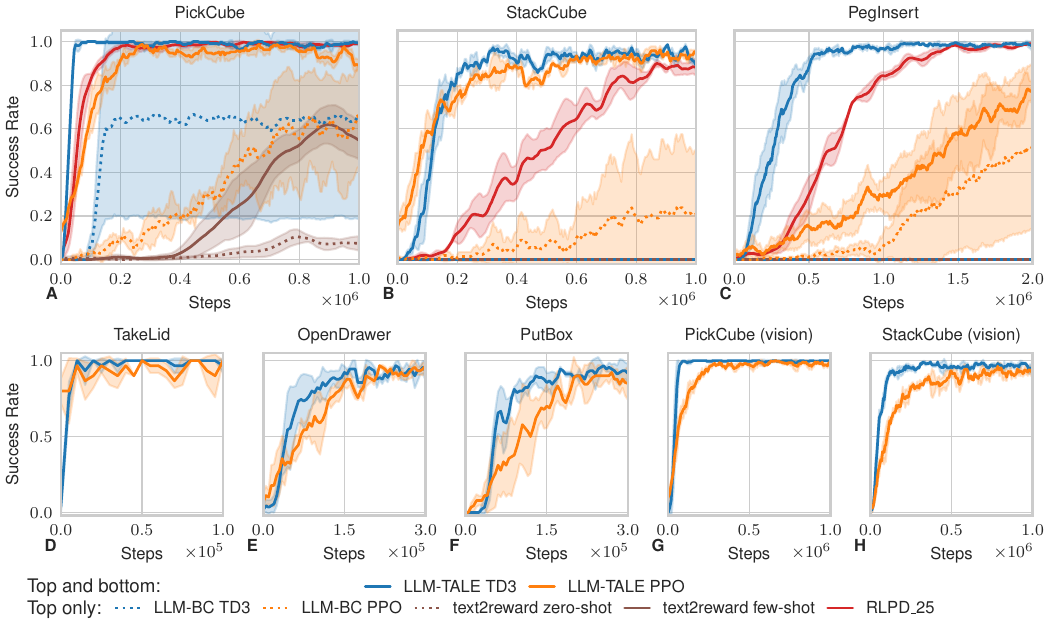}
    \caption{Top figures (\textbf{A-C}) show evaluation results with comparisons against baselines and LLM-TALE ablations in ManiSkill\cite{mu2021maniskill} tasks, while bottom figures show evaluation results in RLBench~\cite{james2020rlbench} (\textbf{D-F}) and vision-based (\textbf{G-H}) tasks.}
    \label{fig:eval}
\end{figure*}

\begin{figure*}
    \centering
    \includegraphics[width=\linewidth]{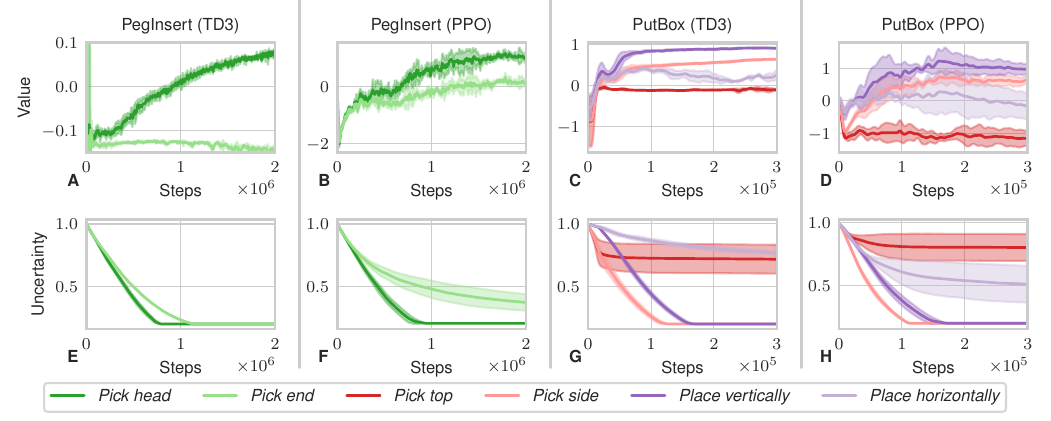}
    \caption{Visualization of value- and uncertainty-based affordance exploration with LLM-TALE for \taskname{PegInsert} and \taskname{PutBox}.}
    \label{fig:exploration}
\end{figure*}

\section{Simulation Experiments}

Our simulation evaluations span six tasks: three from RLBench~\cite{james2020rlbench} and three from ManiSkill~\cite{mu2021maniskill}\footnote{The corresponding code, hyperparameters and prompts are available at \url{https://github.com/llm-tale/llm_tale}.}.
The robot end-effector is controlled with relative position or velocity commands rather than joint-space control, which simplifies the action space for language models.
The action is \(a = (\delta_x, \delta_y, \delta_z, \delta_{rx}, \delta_{ry}, \delta_{rz}, \text{grip})\) and comprises translational and rotational deltas plus a gripper action.
All experiments were repeated three times with different seeds.

We learn policies with a residual action space on top of base actions defined by pre-existing primitives (Alg.~\ref{alg:llm-tale}).
These primitives are implemented with a PD controller that moves linearly toward the target.
The proportional gain is set to 1 by default, and motion is limited by maximum velocity and orientation-velocity bounds.
For the \emph{pick} primitive, the episode terminates when the object’s displacement from its initial position exceeds a threshold.
For the \emph{transport} primitive, termination occurs when the positional error to the goal is below a threshold and the object is nearly static.

We use general dense intrinsic reward functions that apply across the primitive per benchmark suite.
For example, for pick actions in the RLBench suite, the intrinsic reward is
\begin{equation}
r^\mathrm{in}_j \;=\; -\,\tanh\!\bigl(5\, e_\mathrm{pos}\bigr)\;-\;0.25\,\tanh\!\left(\frac{\lVert \dot{q}\rVert}{\pi}\right).
\end{equation}
Here, \(e_\mathrm{pos}\) is the Euclidean distance between the end-effector and the goal position, and \(\dot{q}\) denotes the robot joint velocities.
In addition to the intrinsic reward \(r^\mathrm{in}\), the environment provides an external reward \(r^\mathrm{ex}\) (with \(r^\mathrm{ex} \gg r^\mathrm{in}\)) upon successful task completion or other termination conditions.

We use OpenAI GPT-4o \cite{openai_gpt4o_system_card_2024} for planning, selected for its performance, cost, and latency.
Each planner prompt includes a brief task description, specific instructions, and task-related knowledge.
We also provide a few examples to specify the desired format and reasoning.
Consistent with Alg.~\ref{alg:llm-tale}, we query the LLM before training and cache the outputs as Python functions and dictionaries for reuse during training.

\subsection{ManiSkill Tasks}
We evaluate the LLM-TALE framework on three representative pick-and-place tasks from the ManiSkill suite~\cite{mu2021maniskill}: \taskname{PickCube}, \taskname{StackCube}, and \taskname{PegInsert} (short for \taskname{PegInsertionSide}).
These tasks are shown in Fig.~\ref{fig:tasks}.
This experiment evaluates sample efficiency by comparing our method against three baselines and an ablation of LLM-TALE: Text2Reward~\cite{xie2023text2reward} (zero-shot and few-shot), RLPD~\cite{ball2023efficient} with 25 high-quality demonstrations, and LLM-BC (ablation of LLM-TALE).

\noindent\textbf{Text2Reward} uses LLM reasoning to generate dense rewards, guiding robots through task steps and fostering meaningful behavior without expert reward examples or human fine-tuning~\cite{xie2023text2reward}.
There are two settings: zero-shot and few-shot.
Zero-shot: the LLM sees only the task description and generic instructions (no task-specific examples).
Few-shot: the prompt adds one example reward function to guide reward generation~\footnote{Text2Reward is run with the authors’ original ManiSkill2-based code, whereas all other results use ManiSkill3.}.

\noindent\textbf{RLPD} is an efficient RL method that leverages demonstrations, using \(50\%\) offline data and 10 critics to prevent overfitting.
We use the implementation from \cite{tao2024rfcl} and train with sparse rewards and 25 demonstrations.

\noindent\textbf{LLM-BC} is an ablation of LLM-TALE that replaces the PD controller with a behavior cloning (BC) base policy.
This policy is trained on offline demonstrations generated by the PD controller following LLM-generated plans.

Fig.~\ref{fig:eval}~A--C show the evaluation success rates for all methods.
Our method directly guides the robot to LLM-generated goals (Alg.~\ref{alg:llm-tale}), yielding fast and stable convergence across all tasks with PPO~\cite{schulman2017proximal} and TD3~\cite{fujimoto2018addressing}.
The only exception is the PPO variant for the \taskname{PegInsert} task, which shows a positive trend yet underperforms the TD3 variant and RLPD\_25.
Although on-policy methods typically have lower sample efficiency than off-policy methods, this gap is reduced in our setting.
This can be attributed to primitives that move toward LLM-generated goals, which provide effective actions that simplify policy optimization.
Compared with RLPD\_25 and Text2Reward, our approach is more sample-efficient in most settings.
A further advantage over RLPD is that LLM-TALE does not require (human) demonstrations.

In the LLM-BC ablation, the training strategy (rewards and exploration) is identical; only the base policy differs.
The TD3 version achieves some success on \taskname{PickCube} but shows no improvement on the other tasks.
The PPO on-policy variant shows a positive trend for all tasks but underperforms LLM-TALE.
The BC base policy fails under out-of-distribution states, which conflicts with exploration during RL training and makes it less efficient than a stable PD-based policy.

We also evaluate vision-based variants on two tasks (Figs.~\ref{fig:eval}G--H) to validate that our method can handle high-dimensional observations.
Following DrQ-v2~\cite{yarats2021mastering}, we use a similar encoder with random-shift augmentation and a CNN backbone for visual encoding.
This setup attains high sample efficiency in on- and off-policy learning, comparable to state-based input.

\begin{table}[t]
\centering
\caption{Tasks with multiple identified affordance modalities ($m>1$). The remaining tasks have $m=1$.}
\label{tab:multimodality}

\renewcommand{\arraystretch}{0.95}
\setlength{\tabcolsep}{5pt}

\begin{tabularx}{\linewidth}{l c X}
\toprule
\textbf{Task} & $m$ & \textbf{Affordance Description} \\
\midrule
\taskname{PegInsert} & 2 & Pick the peg from the end or head. \\
\taskname{PutBox}    & 4 & Pick the box from the side or top; place vertically or horizontally in the cupboard. \\
\bottomrule
\end{tabularx}
\end{table}

\subsection{RLBench Tasks}
RLBench~\cite{james2020rlbench} provides a wide range of everyday tasks, making it a suitable platform for evaluating our LLM task planning and affordance-level planning pipeline.
We evaluate LLM-TALE on three additional tasks: \taskname{TakeLid}, \taskname{OpenDrawer}, and our motivating task \taskname{PutBox}.
These tasks are also shown in Fig.~\ref{fig:tasks}.
We use the control mode \taskname{EndEffectorPoseViaIK}, which issues relative position and orientation commands.

The \taskname{PutBox} task was implemented in RLBench by modifying the existing \taskname{PutGroceriesInCupboard} task.
We remove other objects from the original scene and add a constraint that the robot must not tilt the box to avoid spilling its contents.
This constraint is not specified to the LLM during planning; instead, we incorporate it as an external reward during RL training.
We also include collision detection as an external penalty signal.
The method explores both affordance and execution levels to learn a policy that places the box in the cupboard.

Results for these tasks are shown in Fig.~\ref{fig:eval}~D--F and indicate high sample efficiency for TD3 and PPO variants.

\subsection{Affordance Exploration}

Beyond task performance, we evaluate the affordance exploration mechanism of LLM-TALE.
Table~\ref{tab:multimodality} reports the outputs of the affordance-modality identifier.
With the affordance identifier prompt \(\mathcal{M}\), the LLM finds multimodal affordances for \taskname{PegInsert} and \taskname{PutBox}.
For \taskname{PegInsert}, the LLM-generated plan includes picking the peg from the head or from the end.
For \taskname{PutBox}, the identifier returns top and side picks and horizontal or vertical placements.

Affordance-exploration results are visualized in Fig.~\ref{fig:exploration}.
In Fig.~\ref{fig:exploration}~A, the TD3 and PPO variants quickly learn that head grasps yield higher value than end grasps.
Fig.~\ref{fig:exploration}~E shows that the agent still explores end grasps, but less frequently, as their uncertainty decays more slowly than for head grasps.
The PPO variant shows similar trends but takes longer to separate the modalities.
For \taskname{PutBox}, both TD3 and PPO eventually stop exploring top picks, since this choice hampers accurate placement without spillage (Fig.~\ref{fig:exploration}~G--H).
Figs.~\ref{fig:exploration}~C--D show that both variants identify side picks and vertical placements as the highest-value modalities after approximately 50k steps.

\section{Real-World Experiments}
To assess real-world performance, we evaluate zero-shot transfer of a simulation-trained policy on the motivating task \taskname{PutBox} using a physical setup.
In this experiment, we compare LLM-TALE with the LLM-generated linear base policy to evaluate the effectiveness of our residual learning approach.

\subsection{Experimental Setup}
\begin{figure}[t]
    \centering
    \begin{subfigure}[b]{0.9\columnwidth}
        \centering
        \includegraphics[width=0.85\textwidth]{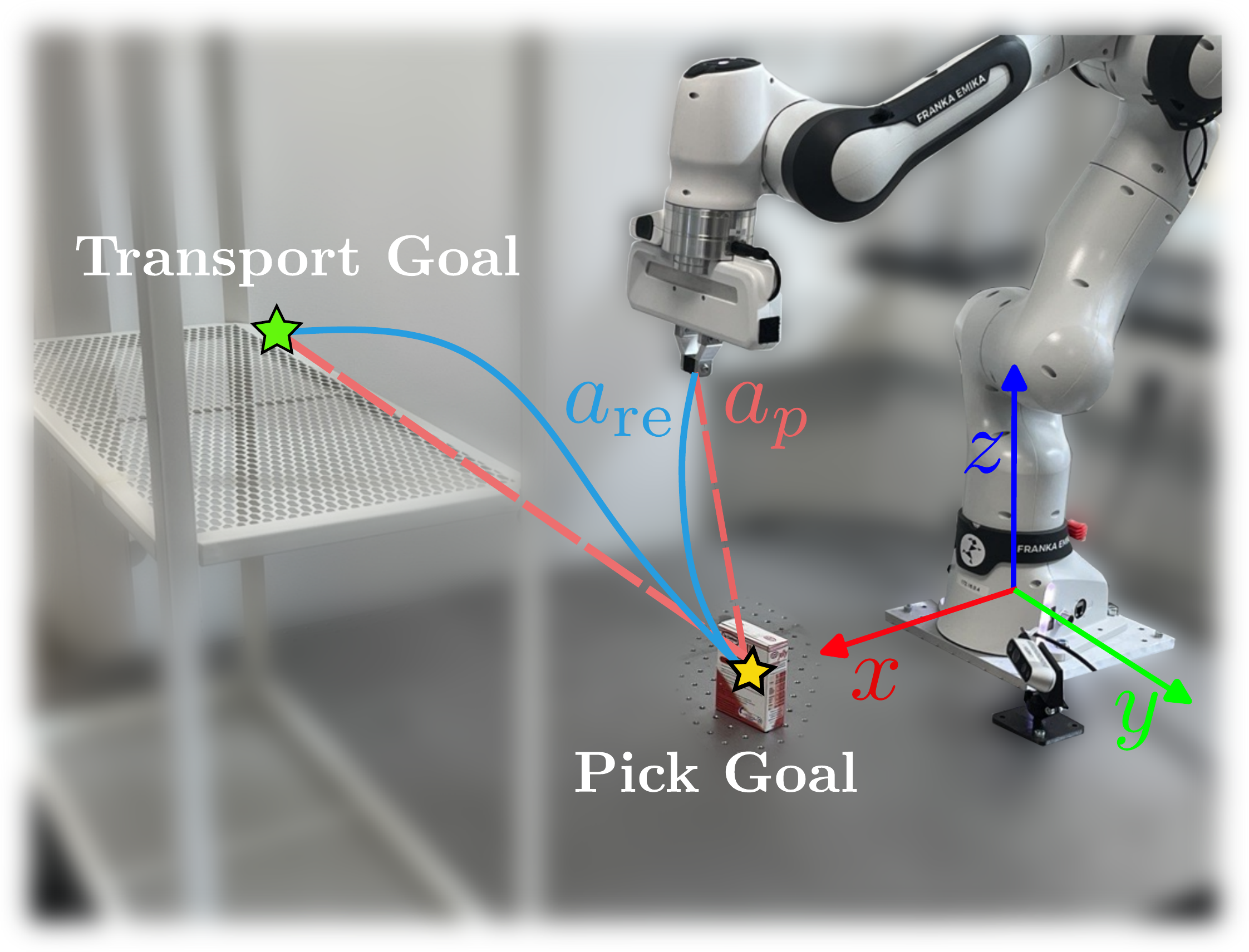}
        \caption{Real-world \taskname{PutBox} setup.}
        \label{fig:real_robot_overview}
    \end{subfigure}
    \hfill
    \begin{subfigure}[b]{0.45\columnwidth}
        \centering
        \includegraphics[width=0.95\textwidth]{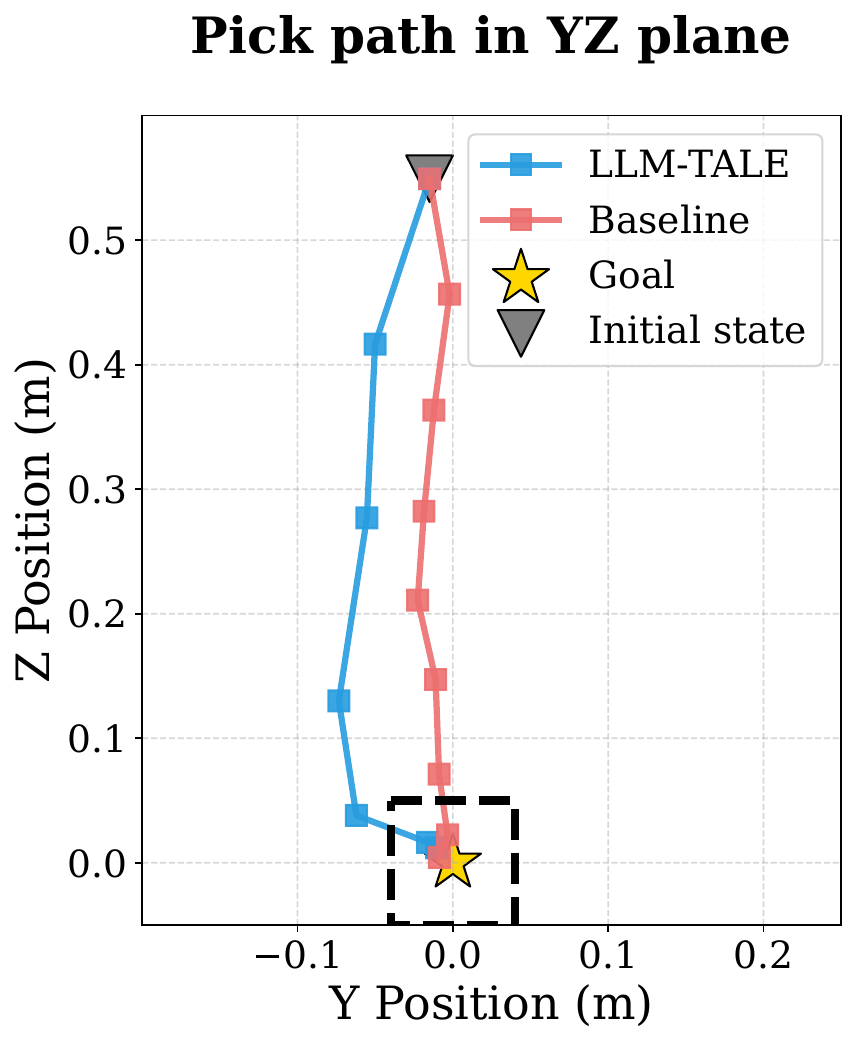}
    \end{subfigure}
    \hfill
    \begin{subfigure}[b]{0.48\columnwidth}
        \centering
        \includegraphics[width=0.95\textwidth]{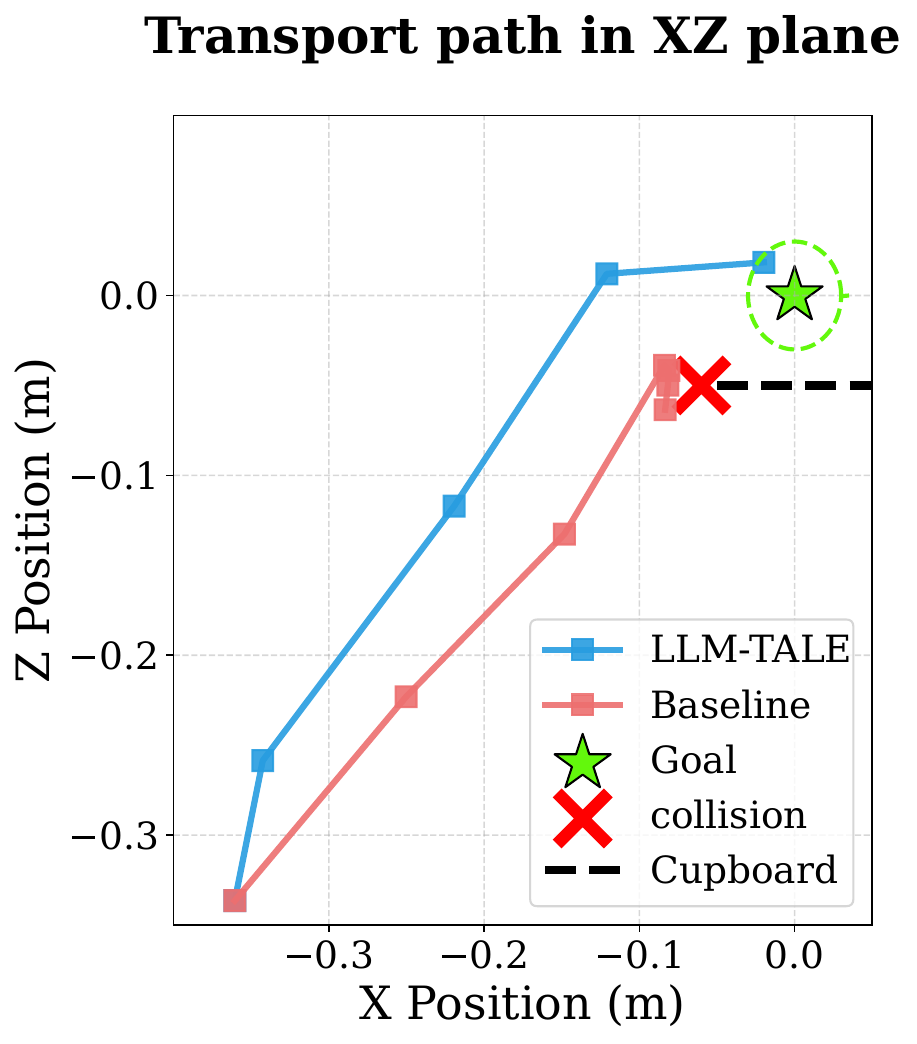}
    \end{subfigure}
    \caption{Zero-shot sim-to-real experiment for the \taskname{PutBox} task.}
    \label{fig:real_robot_experiment}
\end{figure}

We used a Franka Emika Panda with a Franka gripper, controlled by a Cartesian impedance controller that commands the end-effector pose at 1 kHz.
Our method sent end-effector pose commands to the impedance controller at a lower rate.
We performed real-time object tracking using a RealSense D435 RGB-D camera, as shown in Fig.~\ref{fig:real_robot_overview}.

\subsection{Results}
We deployed a policy trained in simulation using LLM-TALE with TD3 on the robot and ran 15 episodes using our method and an LLM-only controller that directly executed LLM-generated plans.
Our method achieved a success rate of \(93.3\%\) with one failure, while the LLM-only base policy had a \(0\%\) success rate due to collisions with the cupboard.

Figs.~\ref{fig:real_robot_experiment}~A and \ref{fig:real_robot_experiment}~B show trajectories used to complete the task.
The base primitive policy produces linear actions directed toward the goal pose identified by the LLM planner.
Even without specific sim-to-real techniques, the primitive-based policy guides the robot toward higher-confidence regions when encountering unknown states.

The residual policy refines trajectories to ensure physical feasibility.
The trajectory in Fig.~\ref{fig:real_robot_experiment}~A shows that the RL agent learns to avoid collisions during picking.
Moreover, the trajectory in Fig.~\ref{fig:real_robot_experiment}~B shows that the residual policy increases vertical clearance during placement to avoid contact with the cupboard.
The robot also executes larger, steadier motions, completing the task more efficiently than the LLM-only controller.
Overall, the residual policy outperforms the LLM-only policy in both success rate and reliability.

\section{Conclusion and Discussion}
In this paper we introduce LLM-TALE, a framework for LLM-guided exploration at the task and affordance levels.
We evaluate the approach in simulation against baselines based on LLM-guided reward shaping and sample-efficient reinforcement learning from demonstrations.
LLM-TALE generates successful affordance-level plans, identifies multimodal solutions, and improves both success rate and sample efficiency over the baselines, even when only a single modality is discovered.
The method is particularly suited for tasks with multiple affordances where the LLM lacks physical understanding.
We demonstrate this in the motivating task \taskname{PutBox}, where the agent initially picks the box from the top but later learns to grasp it from the side, enabling successful placement without spillage.
Real-world evaluations further demonstrate promising sim-to-real transfer.

Despite these advantages, the current planning framework does not handle objects with complex geometry and is limited to simpler items such as cubes, boxes, and drawer handles.
The method also requires access to object state, which requires state estimators for real-world deployment.

To address these limitations, we plan to incorporate interactive learning in which a human provides one or a few demonstrations to help the robot plan for objects with more complex geometric relationships.
To improve generalization, we also aim to use manipulation foundation models, such as AnyGrasp~\cite{fang2023anygrasp}, to generate affordance goals without requiring precise object-state information.

\section{Acknowledgments}

Research reported in this work was partially or completely facilitated by computational resources and support of the Delft AI Cluster (DAIC) \cite{DAIC} at TU Delft (RRID: SCR\_025091), but remains the sole responsibility of the authors, not the DAIC team.

\bibliographystyle{jabbrv_IEEEtran}
\bibliography{bib}

\end{document}